\documentclass[twocolumn,10pt]{asme2e}

\usepackage{graphicx}
\usepackage{svg}
\usepackage{times}
\usepackage{soul}
\usepackage{url}
\usepackage[hidelinks]{hyperref}
\usepackage[utf8]{inputenc}
\usepackage[small]{caption}
\usepackage{graphicx}
\usepackage{amsmath}
\usepackage{amsthm}
\usepackage{booktabs}
\usepackage{algorithm}
\usepackage{algorithmic}
\usepackage{bbm}
\usepackage{amsmath}
\usepackage{amssymb}
\usepackage[inline]{enumitem}

\newcommand{\etal}{{\em et~al.}}
\newcommand{\ie}{{\em i.e.}}
\newcommand{\etc}{{\em etc.}}
\newcommand{\RNum}[1]{\uppercase\expandafter{\romannumeral #1\relax}}
\confshortname{IDETC/CIE 2021}
\conffullname{the ASME 2021\\ International Design Engineering Technical Conferences and\\Computers and Information in Engineering Conference}

\confdate{August 17-20}
\confyear{2021}
\confcity{Virtual}
\confcountry{Online}

\papernum{DETC2021-XXXX}

\title{CreativeGAN: EDITING GENERATIVE ADVERSARIAL NETWORKS FOR CREATIVE DESIGN SYNTHESIS}
\author{Amin Heyrani Nobari\thanks{Address all correspondence to this author.} 
    \affiliation{Dept. of Mechanical Eng.\\
	Massachusetts Institute of Technology\\
	Cambridge, Massachusetts, 02139\\
	ahnobari@mit.edu
    }
}

\author{Muhammad Fathy Rashad
    \affiliation{
	Dept. of Electrical and Electronic Eng.\\
	Universiti Teknologi PETRONAS\\
	Seri Iskandar, Perak, Malaysia, 32610\\
    muhammad.fathy\_25547@utp.edu.my
    }	
}

\author{Faez Ahmed
    \affiliation{Dept. of Mechanical Eng.\\
	Massachusetts Institute of Technology\\
	Cambridge, Massachusetts, 02139\\
	faez@mit.edu
    }
}

\begin{document}

\maketitle    

%%%%%%%%%%%%%%%%%%%%%%%%%%%%%%%%%%%%%%%%%%%%%%%%%%%%%%%%%%%%%%%%%%%%%%
\begin{abstract}
Modern machine learning techniques, such as deep neural networks, are transforming many disciplines ranging from image recognition to language understanding, by uncovering patterns in big data and making accurate predictions. They have also shown promising results for synthesizing new designs, which is crucial for creating products and enabling innovation. Generative models, including generative adversarial networks (GANs), have proven to be effective for design synthesis with applications ranging from product design to metamaterial design.
These automated computational design methods can support human designers, who typically create designs by a time-consuming process of iteratively exploring ideas using experience and heuristics. 
However, there are still challenges remaining in automatically synthesizing `creative' designs. GAN models, however, are not capable of generating unique designs, a key to innovation and a major gap in AI-based design automation applications. This paper proposes an automated method, named CreativeGAN, for generating novel designs. It does so by identifying components that make a design unique and modifying a GAN model such that it becomes more likely to generate designs with identified unique components. The method combines state-of-art novelty detection, segmentation, novelty localization, rewriting, and generative models for creative design synthesis. Using a dataset of bicycle designs, we demonstrate that the method can create new bicycle designs with unique frames and handles, and generalize rare novelties to a broad set of designs. Our automated method requires no human intervention and demonstrates a way to rethink creative design synthesis and exploration.
%Automated design synthesis is key to design automation. Artificial intelligence and machine learning have transformed many domains of design have been successfully automated using data-driven approaches. Modern generative models, including Generative Adversarial Networks, have proven to be effective for design synthesis with applications ranging from product design to metamaterial design. These models, however, are not capable of generating unique designs, a key to innovation and a major gap in AI-based design automation applications. This paper proposes an automated method, named CreativeGAN, for generating novel designs by modifying an existing GAN model such that it becomes more likely to generate designs with components similar to existing unique designs. The method combines state-of-art novelty detection, segmentation, novelty localization, rewriting, and generative models for creative design synthesis. Using a dataset of bicycle designs, we demonstrate that the method can create new bicycle designs with unique frames and handles, and generalize rare novelties to a broad set of designs. Our automated method requires no human intervention and demonstrates a way to rethink creative design synthesis and exploration.
\end{abstract}

%%%%%%%%%%%%%%%%%%%%%%%%%%%%%%%%%%%%%%%%%%%%%%%%%%%%%%%%%%%%%%%%%%%%%%
% \begin{nomenclature}
% % \entry{A}{You may include nomenclature here.}
% % \entry{$\alpha$}{There are two arguments for each entry of the nomemclature environment, the symbol and the definition.}
% if necessary
% \end{nomenclature}
%%%%%%%%%%%%%%%%%%%%%%%%%%%%%%%%%%%%%%%%%%%%%%%%%%%%%%%%%%%%%%%%%%%%%%
\vskip -0.2in
\section{INTRODUCTION}
In engineering design synthesis creativity and innovation are among designers' and engineers' most important objectives. Creativity in design has been a major topic of research; such creativity has been correlated with design and designers' success ~\cite{creativity_assessment}. Design synthesis tools must address this aspect of the design process. 

Creativity is not easily defined, despite many attempts to do so. Sarkar~\etal, explore the literature surrounding the topic of creativity and definitions of creativity and propose their ``common" definition of creativity in design as having ``novelty" and ``usefulness"~\cite{SARKAR2011348}. The synthesis of ``useful" designs is often correlated to their quality ~\cite{SHAH2003111}, and their ``usefulness" is, therefore, measured based on quality. Engineering design tools are usually built with quality, hence usefulness, in mind and, therefore, focus primarily on this aspect of the process but seldom address novelty. In this paper, we focus our efforts on data-driven methods for design synthesis and propose an approach for guiding existing generative models to synthesize novel designs.

The conventional engineering design process involves iteratively and often manually exploring design alternatives and ideas. Designers must spend significant time examining numerous design alternatives until they identify useful solutions. This involves repeatedly evaluating all options, often using physics-based simulations, and adapting designs based on their quality and the design program's requirements. To overcome this, the engineering design community has made significant strides in speeding up or eliminating the conventional iterative design cycle. Topology ~\cite{duysinx1998topology,bendsoe2013topology} and adjoint-based optimization~~\cite{anderson1999aerodynamic} are examples of this effort commonly used in structural design and aerodynamic design, respectively. These methods automate the design synthesis process based on performance requirements, but they still be time-consuming due to the high computational cost of evaluations. 

To overcome this problem, data-driven methods such as generative adversarial networks (GANs)~\cite{GAN} and variational auto-encoders (VAEs)~\cite{VAE}, have been employed in many design synthesis problems~\cite{padgan,mo-padgan,yilmaz2020conditional,achour2020development,Oh_2019,jang2021generative,yoo2021integrating}. GANs and VAEs are generally capable of learning complex distributions of existing designs and even considering performance and quality evaluation when generating new designs~\cite{padgan,mo-padgan}. They allow for learning an underlying low-dimensional latent space that can represent the existing designs. These data-driven approaches allow designers to rapidly generate new designs and reduce the complexity of design space exploration, greatly improving its efficiency. These methods, however, are lacking any mechanism for promoting novelty, a typical lack in most design automation tools.

In this paper we focus on using GANs for design synthesis.
%GANs can be difficult to train, and they suffer from problems that may significantly hinder effective design synthesis. 
One major problem of a typical GAN is \emph{mode collapse}, which refers to the phenomena in which the GAN model learns to generate samples from one or few modes of the design space but misses many other modes~\cite{improvinggan}. 
This typically occurs in more common modes of the data, meaning the GAN will learn to produce only the most plausible output, leaving GANs inherently an inadequate method when it comes to novelty. 
To ameliorate this problem, researchers have proposed methods, such as reconstruction networks proposed by Srivastava~\etal, in VEEGAN~\cite{VEEGAN} or DPP-based promotion of diversity in generated samples proposed by Chen~\etal, in PaDGAN. Although these methods help with mode collapse, the nature of GANs methods will always tend towards data used to train them. This ``emulative''\cite{can} property of GANs is understood well, and it has been shown that GANs will rarely produce designs outside the original distribution of data~\cite{mo-padgan,padgan,can} In addition to those rare designs, they can also produce designs that interpolate existing data and fill large gaps in the design space not filled by existing designs\cite{mo-padgan,padgan}. The rarity of such GAN-generated samples means they are not very useful and will likely be ignored. The current state of the art in GANs lacks mechanisms for generating novel samples.
 
Recent developments in GANs have proven they are useful for generating realistic designs and in fact, can be adapted to generate high quality~(\ie, useful) designs. Novelty, however, is a significant aspect of creativity. In fact, novelty is often emphasized in design creativity~\cite{SARKAR2011348}. Data-driven methods, such as GANs, are capable of generating many candidates, but very few novel designs. 
If these existing models could be modified to create novel designs, they would allow for the development of creative automated design synthesis tools. In this paper we propose \emph{CreativeGAN} --- an approach to modify GAN models to synthesize novel designs. Our primary objective is to bridge the creativity gap in GAN-based design synthesis approaches by promoting novelty in GAN-generated designs. We combine state of the art StyleGAN2~\cite{stylegan2ada,stylegan2,stylegan}, which is capable of generating realistic designs with recent developments in \emph{rewriting} GANs~\cite{bau2020rewriting} through automatic detection and localization of novel features in generated designs. We achieve this by employing a K-nearest neighbour~(KNN) approach to detect novel samples and identifying those features that make them novel. We then use these novel features to modify trained StyleGAN2 models such that they would generate samples with the detected novel features. Doing this with many different unique features will allow designers to guide their data-driven models towards different novel designs. We demonstrate the results of our approach through two proof-of-concept examples within bike design application. We also provide two metrics, to quantify how the novel designs differ from designs that are common in the training data. Our main contributions are as follows:
\vskip -0.2in
\begin{enumerate}
    \item We introduce an approach to systematically modify GAN models to synthesize novel designs in an automated fashion without the need for human interaction or involvement. 
    \item Using a bicycle synthesis application, we demonstrate that our method increases the novelty of designs generated by GANs, thereby showing the potential for automating creativity in data-driven design synthesis.
    \item We demonstrate that anomaly detection algorithm when applied at different granularities can identify unique designs and unique components within each design.
    \item We show the efficacy of a deep convolutional neural network~(CNN)-based semantic segmentation model to predict the location of parts of a bicycle for any new design, with an overall intersection over union (IoU) score of 83.8\%.
    % \item We propose a data augmentation approach that leads to significant improvement in the quality of GAN-generated designs by combining individual parts with assembly data.
    \item We provide two metrics for measuring novelty in GAN-generated samples, one using image structural similarity and the other combining features from a pre-trained neural network with the nearest-neighbor approach.
\end{enumerate}
\vskip -0.2in
\section{RELATED WORKS}
In this section, we provide a brief background on related topics seminal to this work. In this paper, we combine GANs and novelty detection methods to guide novelty and creativity in GANs, we will discuss these topics briefly here. If interested, readers are encouraged to refer to the sources cited for a more in-depth coverage of these topics.

\subsection{Generative Adversarial Networks}
A generative adversarial network~\cite{GAN} is usually made up of two models~\textemdash~a \textit{generator} and a \textit{discriminator}. The generator $G$ maps arbitrary noise distribution to the data distribution, the discriminator $D$ tries to to distinguish between real and generated data. As $D$ improves its classification ability, $G$ also improves its ability to generate data that fools $D$. Based on this GANs have th following objective:
\begin{equation}
\begin{split}
\min_{G} \max_{V} V(D, G)= & \mathbb{E}_{x \sim p_{\text {data}}(x)}[\log{D(x)}]+\\ & \mathbb{E}_{x \sim p_{z}(z)}[\log{(1-D(G(z)))}],
\end{split}
\label{eqn:1}
\end{equation}
where $p_{\text{data}}$ refers to the distribution of data and $p_{\text{z}}$ refers to the distribution of noise. In this approach for training, the generator will learn to relate the noise distribution $p_{\text{z}}$ to the design space distribution of the data $p_{\text{data}}$. GANs are notoriously difficult to train and may often be unstable~\cite{improvinggan}, therefore it may be unreasonable to use this approach for generating realistic designs. Recent developments in this area have improved GANs significantly and established a new state of the art in the quality of GAN- generated samples. We use the state-of-the-art GAN in single class image generation called StyleGAN2~\cite{stylegan2ada,stylegan2,stylegan}, which is capable of generating realistic and high quality images. Further, the research in the design community has proven the efficacy of GAN based automated design synthesis in recent years~\cite{yoo2021integrating,jang2021generative,Oh_2019}, which makes GANs a suitable option for data-driven automated design synthesis.

\subsection{The Creativity Problem In GANs} 
The objective of GANs (Eq.~\ref{eqn:1}) encourages the generator to fool the discriminator, while the discriminator learns the distribution of the data. In this way the generator is ultimately learning to mimic the data, which makes GANs ``emulative"~\cite{can}. This ``emulative" nature has motivated researchers to look into areas where GANs can be pushed towards creativity. Elgammal~\etal, investigate this matter in their Creative Adversarial Network~(CAN)~\cite{can}. If GANs are pushed too far from the original data distribution, their generated samples will lose quality and meaning. For design synthesis, this would mean loss of usefulness in generated designs. Therefore, Elgammal~\etal suggest that the  generator in GAN cannot be pushed too far from original distribution of data~\cite{can}. They, therefore take the approach of maximizing entropy in the generated samples. Others have shown that simply promoting diversity in GANs will allow GANs to fill in data distribution gaps and even deviate slightly from the original distribution\cite{mo-padgan,padgan}. In this work we take a different approach and promote self creativity in GANs. We do this by identifying instances when GANs are being creative, however rare this may be, and guide the generator towards these creative approaches. In this way, we utilize the GANs' own instances of novelty and creativity. %By doing this we can rely on the GAN objective for usefulness in designs without having to worry about a new loss terms leading to less realistic, hence less useful, designs.

\subsection{Detecting Novelty}
In this section, we discuss methods to detect and score novel designs in a dataset. This type of novelty detection is often investigated as a form of anomaly detection~\cite{sabokrou2018adversarially}. In this paper we will also be focusing on this approach as well. In anomaly detection for image data, recent literature on the subject has shifted focus significantly more towards data-driven, specifically machine learning, approaches~\cite{sabokrou2018adversarially,SPADE,10.1145/3097983.3098052,10.1007/978-3-319-59050-9_12,dimattia2019survey,akcay2018ganomaly,zenati2019efficient}. Machine learning based approaches have been proven effective in detecting anomaly/novelty in images, with some researchers using generative adversarial networks to detect novelty by projecting images to the latent space of GANs and measuring the difference between GAN reconstructed images and the actual images~\cite{dimattia2019survey,akcay2018ganomaly,zenati2019efficient}. Others have introduced similar approaches in variational auto-encoders~(VAEs), where instead of reconstructing images using GANs they do so using VAEs~\cite{Xu_2018}. Others use classifiers and non-generative models to detect novelty effectively by using intermediate features of classifiers or training classifiers to identify novel samples~\cite{sabokrou2018adversarially,SPADE}. In this paper we are also interested in understanding what features make images novel. In this paper we use an approach which requires no training and utilizes intermediate features of pre-trained classifiers to detect novelty and localize novel features called, `Semantic Pyramid Anomaly Detection (SPADE)'~\cite{SPADE}.

\section{BACKGROUND}
Here we will discuss the details of some of the approaches in literature that we have implemented in our methodology. We use the SPADE~\cite{SPADE} approach for detecting and localizing novelty in designs and use the rewriting GANs method proposed by Bau~\etal, to guide GANs towards novel features~\cite{bau2020rewriting}. We briefly discuss some of the details of these approaches in this section, however readers  are  encouraged  to refer to the original sources for a more in-depth understanding of these topics.
\subsection{Rewriting GANs}
GANs trained on existing designs are capable of learning the physical and semantic rules in the datasets. Once the training has been completed, however, a mechanism is needed to alter these rules to synthesize designs which are novel. This was made possible recently thanks to the approach introduced by Bau~\etal, called \emph{rewriting GANs}~\cite{bau2020rewriting}. In their approach, the authors propose a generalizable method for editing GANs based on examples of desired changes. In the case of images, these changes may be made by adding a new feature to existing images. For such a change to be possible, the generator $G$ must be altered to incorporate this desirable behavior. The naive way of doing this would be to retrain the model using edited images. This would present two challenges: all images with features we desire to alter must be manually edited, and training large GAN models can be inefficient. Bau~\etal, address these challenges.

Given a trained generator $G\left(z ; \theta_{0}\right)$ with weights $\theta_{0}$, samples designs $x_{i}=G\left(z_{i} ; \theta_{0}\right)$, can be synthesized based on a latent code $z_{i}$. At this point we would have desired changes that we want to enforce in the generator. Imagine we have manually edited some generated design to include desired changes $x_{* i}$. We would update the generator weights $\theta_{1}$ to produce our desired changes examples $x_{* i} \approx G\left(z_{i} ; \theta_{1}\right)$. An appropriate editing would not interfere with other rules and behaviours in the generator. This can be done using the following objective~\cite{bau2020rewriting}:
\begin{equation}
\theta_{1}=\arg \min _{\theta} \mathcal{L}_{\text {smooth }}(\theta)+\lambda \mathcal{L}_{\text {constraint }}(\theta),
\label{eqn:2}
\end{equation}
\vskip -0.5in
\begin{equation}
\mathcal{L}_{\text {smooth }}(\theta) \triangleq \mathbb{E}_{z}\left[\ell\left(G\left(z ; \theta_{0}\right), G(z ; \theta)\right)\right],
\end{equation}
\vskip -0.5in
\begin{equation}
\mathcal{L}_{\text {constraint }}(\theta) \triangleq \sum_{i} \ell\left(x_{* i}, G\left(z_{i} ; \theta\right)\right),
\end{equation}
In the above equation, $\ell(\cdot)$ is a distance metric measuring the perceptual distance between the two sets of images and $\lambda$ is parameter determining the relative weight between the two loss terms. If we were to apply the optimization described in Eq.~\ref{eqn:2}, we would effectively retrain the model. As $\theta$ has a large number of parameters, if the number of examples were to be small, the generator will quickly over-fit and may not generalize. To overcome the overfitting issue and simultaneously reduce the computational cost of this optimization, Bau~\etal propose that the changes in the generator be limited to only one layer. This decreases the computational cost and allows for a generalizable change in the generator~\cite{bau2020rewriting}. This, however, is still not sufficient and the number of parameters in one layer can be numerous.  To overcome this, Bau~\etal constrain the degrees of freedom during optimization by limiting changes in the weights to specific directions at every step~\cite{bau2020rewriting}~(or what they call `rank'). 
% The editing process will be generalizable even with a small number of examples and will be limited to a small subset of parameters in $\theta$. This allows for the process to become fast and computationally efficient enabling quick and easy editing of the generator. Optimization must be done on the outputs of each layer individually and not on the outputs of the generator model. We can trace desired features back to a specific layer, making it possible to alter other generated samples based on these layer-specific features. 
In this paper we will take a similar approach: basing our editing on novel samples generated by generator and transferring novel features from these samples to other generated samples. Using the approach developed by Bau~\etal. Our work differs in two main ways from ~\cite{bau2020rewriting}. First, in contrast to the manual approach used in ~\cite{bau2020rewriting}, we identify the base image and the feature which needs to be copied in a completely automated way using a combination of novelty detection, novelty localization, and segmentation analysis. Second, our approach focuses on rewriting for novel designs and attributes. Third, our focus is Engineering Design applications, while ~\cite{bau2020rewriting} focused on computer vision applications.

\subsection{Identifying and Localizing Novelty}
\begin{figure*}[ht]
\centering
\includegraphics[width=1.8\columnwidth]{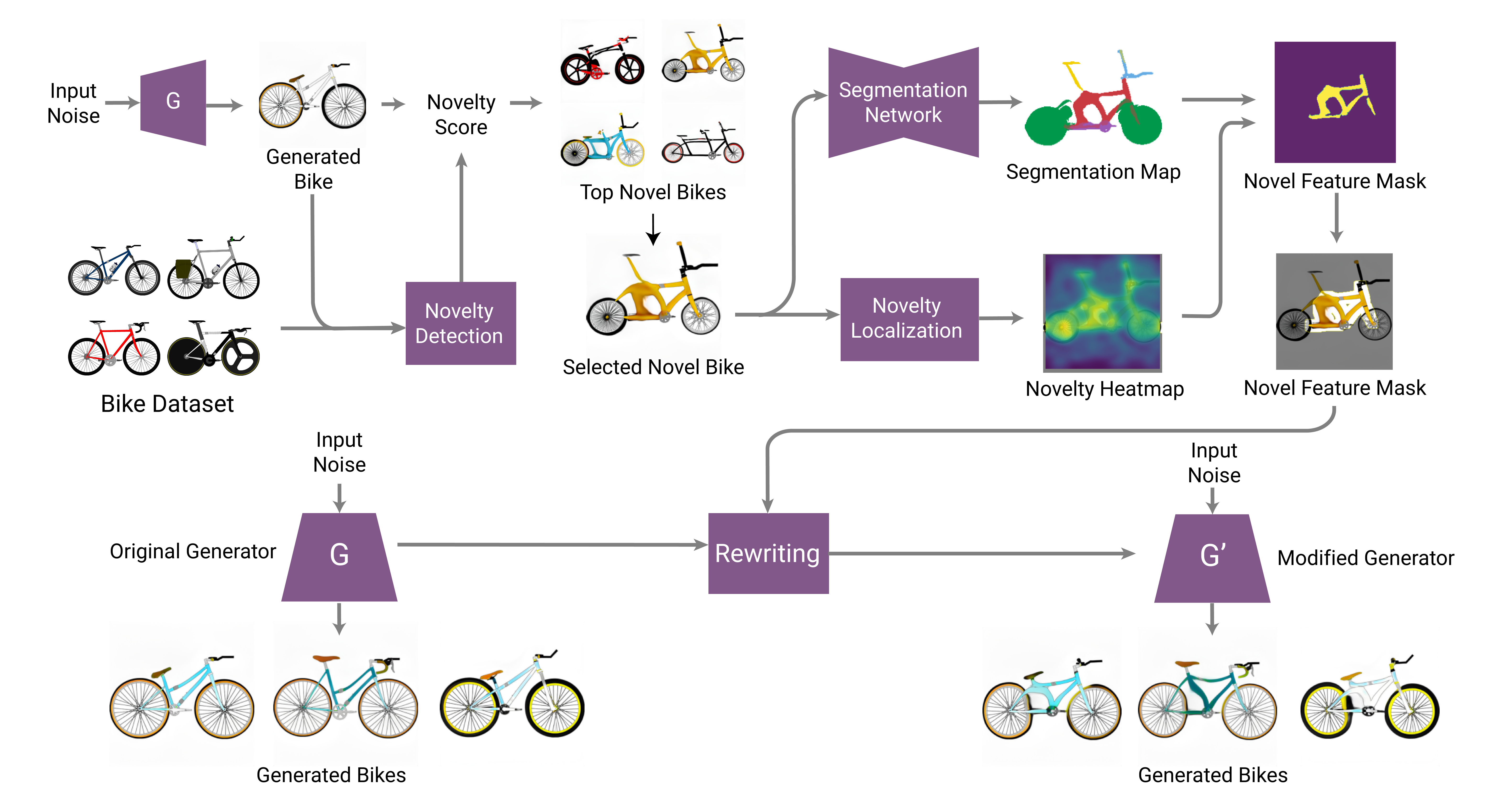}
\caption{Overview of the proposed method used to edit the generator to to generate more novel features.}
\vskip -0.1in
\label{fig:methodoverview}
\end{figure*}
In this paper we intend to modify GAN models to generate novel designs using the rewriting approach. This requires us to identify novel samples and identify the features in any novel design that make that design novel. For this purpose we use the Semantic Pyramid Anomaly Detection (SPADE) method developed by Cohen~\etal~\cite{SPADE}. SPADE is a method based around the K nearest neighbours~(KNN) method, and KNN-based approaches have been shown to be highly effective in novelty/anomaly detection~\cite{MARKOU20032481}. SPADE is more than applying KNN to images. Cohen~\etal, proposed using the global average by pooling the features of the last layer of a ResNet~\cite{resnet} classifier pre-trained on the ImageNet dataset. Bergman~\etal, showed that pre-trained ImageNet features outperform the features learned from self-supervised method ~\cite{bergman2020deep}. This means that the classifier is not trained for the dataset; instead the pre-trained weights of the ResNet model, are used.
To detect image-level novelty~(\ie, determine how novel any generated sample is overall) in SPADE the features from the ResNet $f_{i}$ for all of the existing designs~(\ie, dataset) $x_{i}$ are extracted at the beginning, then during the inference stage, the ResNet features $f_{y}$ of any given sample $y$ is extracted. The novelty score of every image is defined as the mean of the distances between the sample's features $f_{y}$ and the features of the $K$ nearest existing design samples' $N_{k}(f_{y})$:
\vskip -0.1in
\begin{equation}
d(y)=\frac{1}{K} \sum_{f \in N_{K}\left(f_{y}\right)}\left\|f-f_{y}\right\|^{2}
\label{eqn:5}
\end{equation}
\vskip -0.1in
From here the novelty score $d(y)$ of any given sample $y$ can be determined; the higher this value the more novel the sample is expected to be.

Cohen~\etal propose an approach to find the location of the features within an image that make it novel. To do this we again use KNN-based approach, but this time the novelty score is computed for any given pixel in an image rather then the overall image features. For this an approach similar to the computation of overall novelty score will be taken, however, this time the features of multiple layers~(\ie, intermediate features) will be captured for the $K$ nearest samples in the dataset and mapped to the corresponding pixels. Then the novelty score for each pixel location $p$ in a novel sample $y$ will be determined based on the mean of the distances between that pixel's features and the $\kappa$ nearest pixels in the same location amongst the $K$ nearest samples to $y$~(\ie, $N_{k}(f_{y})$):
\begin{equation}
d(y, p)=\frac{1}{\kappa} \sum_{f \in N_{\kappa}(F(y, p))}\|f-F(y, p)\|^{2}
\label{eqn:6}
\end{equation}
where $\kappa \leq K$ and $F(y,p)$ represent the feature extractor for sample $y$ at pixel location $p$. Thus, a map of novelty scores can be determined for any image with a high overall novelty score, and, from these values the locations of novel features can be determined by setting a threshold $\tau$ on the novelty scores of pixels and taking pixel locations with a score higher than the threshold. These locations will form an overall map of novelty in any image, which will determine which of its features are novel.
\section{METHODOLOGY}
In this paper, we show how the approaches described in the previous section can be combined as an automated approach for promoting novelty in GAN-generated designs. We call our method CreativeGAN. CreativeGAN identifies novel designs and the attributes that make a design novel (novel attributes) by combining novelty detection, novelty localization and segmentation algorithms. CreativeGAN then edits a pre-trained GAN to generate samples, which are more likely to exhibit the novel design attributes. CreativeGAN does not just copy and paste novel design features to new designs, instead it integrates those features with other design attributes in the new design. Fig.~\ref{fig:methodoverview} illustrates the overall architecture of CreativeGAN. We showcase our approach through an example of bike design, where we train a GAN to generate images of bikes and guide the generator to produce bikes with more novelty. 
\subsection{Generating Realistic and Useful Designs}
In any GAN-based approach, generated designs should be realistic and useful. We train the state of the art GAN architecture, StyleGAN2~\cite{stylegan2}, which addresses many of these issues, on our dataset of bike designs with each image having 512 pixel x 512 pixel resolution.

\paragraph{Data augmentation:} StyleGAN2, like many other GAN models, often requires a large dataset of images to generate realistic images~\cite{stylegan2ada}. Because our datasets were of limited size, we augmented the size before training StyleGAN2. For augmentation, we employed the adaptive discriminator augmentation (ADA) method proposed by Karras~\etal\cite{stylegan2ada}. In ADA, Karras~\etal propose 6 different differentiable operations that can be done on images to augment the data. Further, they apply these augmentations stochastically with probabilities lower than 1.0, to give non-augmented images dominance so that the transformations do not leak into the generator~\cite{stylegan2ada}. They show that applying this improves the quality of samples generated by stylegan2 when the size of the dataset is small, as is the case here. 
In the interest of space, we do not discuss the augmentation methods any further as it is not central to our key methodology. To increase the quality of our generated samples, we augment the training data by using images of bike parts~(details described in section \ref{sec:data}). The augmentation leads to 27,406 total images in the training dataset. We also find that providing images of bike parts as training data in the early stages of training StyleGAN2 leads to further improvement in the  quality of generated bikes. After training the model with the bike part augmentation, we fine-tune the model with only full bike images to prevent the GAN model from generating partial bikes. Another reason for doing this is to achieve better disentanglement in the StyleGAN2 generator. The separation of the underlying aspects of the design generation, in this case bike parts, in the generator is often referred to as disentanglement, which is important for successful editing of GANs~\cite{bau2020rewriting}. The reason for this is that when editing GANs, if the feature being changed is not disentangled from other features, a change in that feature will cause unwanted and undesirable changes in other features of the images.
\subsection{Detecting and Localizing Novelty in Generated Designs}
\begin{figure*}[ht]
\centering
\includegraphics[width=1.8\columnwidth]{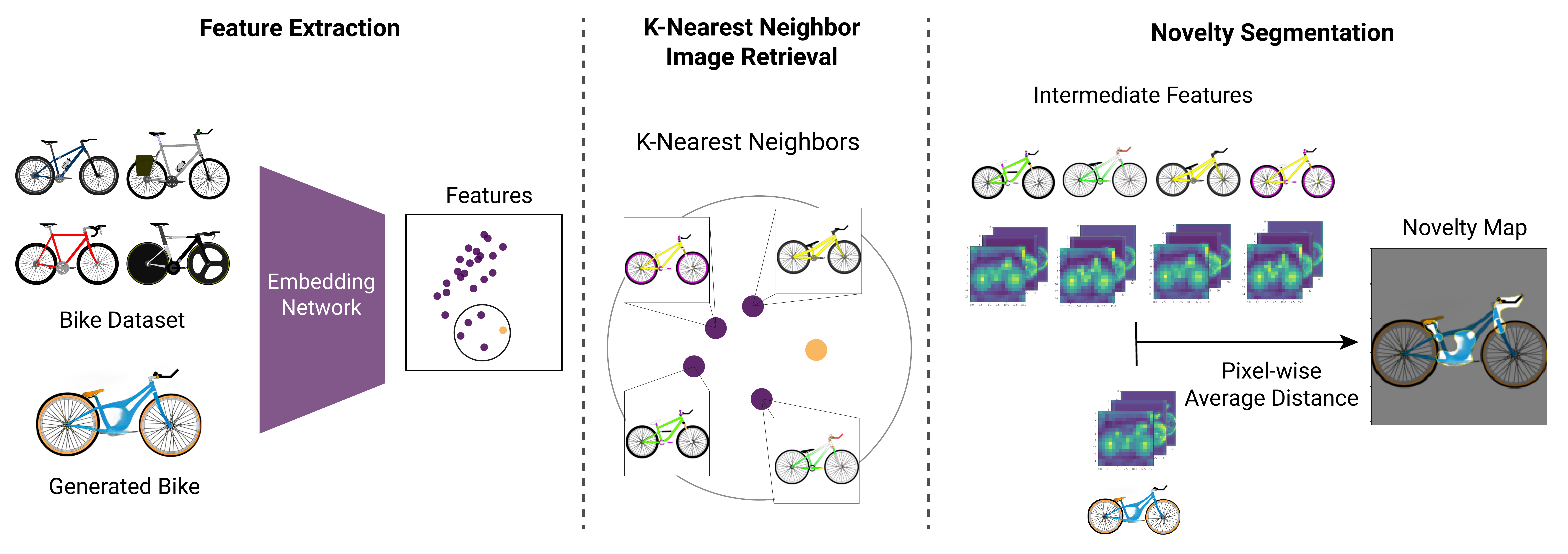}
\vskip -0.1in
\caption{Overview of novelty detection algorithm used to select samples with novel features.}
\label{fig:anomaly}
\end{figure*}
StyleGAN2 generates realistic designs, but some will be more novel. In this section, we discuss how to identify these novel designs and further localize the design attribute in those generated designs that contribute most in making them unique.
Fig.~\ref{fig:anomaly} shows the overall method for detecting and localizing novel designs and their attributes. The process can be divided into three stages: (1)~feature extraction from the dataset images and a test image, (2)~Measuring overall novelty of designs using K-nearest neighbors, and (3)~identifying feature map contributing to novelty. We explain the steps below.
%(3)~computation of overall anomaly score and anomaly map.
\paragraph{Feature extraction for novelty detection:}
%Now that we have established our approach for generating realistic and useful designs we can now identify novel samples and localize their novel features. 
For this purpose, we use the SPADE method described in the prior sections. The first step of detecting novelty using SPADE is to identify which samples exhibit the highest overall novelty. The SPADE method relies on features extracted from deep CNN models rather than the images themselves. In our implementation of SPADE we use the Wide Residual Networks50~(WideResNet50)~\cite{zagoruyko2017wide} architecture pre-trained on the ImageNet dataset. 

To identify and localize novelty in generated designs, we first compute and store the intermediate features of the pre-trained WideResNET50 model as well as the globally averaged features of the final layer of the pre-trained WideResNET50 model. To identify the image-level novelty for any given generated design, we use the globally averaged features of the WideResNET50 model for the samples that are being analyzed. 

\paragraph{Measuring overall novelty using K-nearest neighbors:}
Next, we compute the novelty score for the generated sample using the globally averaged features of the generated sample and the globally averaged features of the 50 nearest neighbors of the generated sample in the dataset as described by Eq.~\ref{eqn:5}. The resultant score is used as a novelty scores of any generated design. 

\paragraph{Identifying feature map contributing to novelty:}
After scoring a set of generated designs on overall novelty and identifying the most novel designs, the next step is to find the features within novel designs that contribute the most to the overall novelty of these designs. To do this, we use a KNN based approach similar to Eq.~\ref{eqn:5}. However, instead of measuring novelty scores for each design, we estimate the novelty of each pixel in each design, and instead of using the globally averaged features, we use the intermediate features of the pre-trained WideResNET50 model.
Using these pixel-wise features and the pixel-wise features of the 50 nearest overall neighbors found in the process of determining the overall novelty score described above, we compute the novelty score for each pixel based on Eq.~\ref{eqn:6}. Then we classify each pixel as belonging to a novel feature if its novelty score exceeds a given threshold. This helps identify a novelty map, highlighting regions within a novel design which are most unique compared to other designs. The rightmost image in Fig~\ref{fig:anomaly} show the novelty feature maps, which help in identifying the area within a design, which is most unique.
%From here we obtain a novelty map determining which regions of the image contribute most towards novelty. % So far, we have established our approach for generating realistic designs and identifying novel items within a set of generated designs. Furthermore, we have proposed an approach for localizing the features that make a generated design novel. Next, we discuss how these identified novel features can help in modifying an existing generative model, so that it synthesizes novel designs.
\subsection{Editing a GAN to Generate Novel Designs}
The original generative model may occasionally create a unique design. However, by identifying what makes the design unique, we can modify the generative model itself to generate many unique designs with similar features. Our goal is to create an approach that can take novel features synthesized by a generative model and rewrite the generalized rules established by the generator towards synthesizing more samples with the novel feature identified. To do this we use the GAN rewriting approach introduced by Bau~\etal~\cite{bau2020rewriting}. Bau~\etal proposed a way to rewrite a GAN model based on a manually identified base image, context images and a mask which needs to be edited. 
 %  Bau~\etal, approached re-writing as a mostly user-centric approach where a user/designer identifies novel features and manually selects how they wish to edit the generator behavior ~(e.g. selecting a smile and editing images without a smile to be smiling manually). 
In this paper, we aim to build an automated design synthesis model which can give designers numerous novel design candidates without any insight or effort needed from the designer. As a person cannot practically sift through thousands of designs to identify novel designs and attributes, selecting novel features and applying them to other generated samples manually. 

The GAN rewriting method relies primarily on three inputs: 
\begin{enumerate*}
  \item a feature that is desired to be emphasized in the generator rules, and 
  \item a primary example of where the desired feature can be transferred to in another generated design, which does not have the desired novel feature.
  \item a secondary set of contextual examples which can help guide the generator to apply the desired features in a contextually appropriate manner. 
  \end{enumerate*}
The contextual and primary examples give context to the model about the real world. By seeing a few examples, the model learns in what context a feature is generally placed in an image. For example, a few context images for a bike handle will allow the model to learn that a bike handle generally appears in the front of a bike and does not hang from the back. Without these contextual examples, there would be no practical way to modify the generator without causing it to generate unrealistic or meaningless samples. 

\paragraph{Automated identification of context:}
To make rewriting possible in an automated fashion for this application, we introduce an approach to not just identify the novel features but also contextually determine where said features could be applied in other common~(\ie, not novel) designs. 
% To automate the process, we leverage the fact that most real-world designs have a well defined known structure of how individual parts are placed relative to each other within an assembly. To do this, we first divide the design assembly (say the design of the bike) into the design of its individual parts or components. Imagine that some part of a bike frame makes a bike more novel. We can assume that, in this context, the entire frame's design is novel. This assumption is supported by the observation that the frame's different portions are connected and function together. 
% % A designer cannot simply change a portion of the frame without rethinking the entire frame design. If they simply copied the feature to parts of their frame, the new design will likely make little to no sense.  
% We use this observation in identifying what part of a design should be used for re-writing GANs. Instead of using a part of image identified in the previous steps, we now identify the bike part which is associated with the specific part.
We do this by identifying which part of a design (of the 7 parts of any given bike) has the most overlap with the novelty feature identified earlier. When applying the rewriting method, we also observed that the generated designs were more realistic when the transfer of features happened from the entirety of parts rather than only a partial segment of parts, which confirmed our intuition.

\paragraph{Predicting design parts:}
A challenge in using parts of a design generated by a GAN for rewriting is that we do not have labels of different parts within an assembly. Hence, we first use a machine learning model to predict the parts of a generated bike.
 We train a segmentation model~(with a U-Net Architecture~\cite{ronneberger2015unet}) on the bike part data we obtained from the original bike dataset as described earlier~(Fig.~\ref{fig:segmentation_data}). This model allows us to identify bike parts in generated samples where the segmentation information is not available. At this point to identify the novel \emph{part} of the bike rather than the localized novel feature, we combine the novelty feature and the segmentation masks of different parts to determine which part of the bike, the novel feature belongs to. To do this we compute the intersection over union~(IoU) of each part of the bike and the novelty map and choose the part of the bike with the highest IoU as the novel \emph{part} of that bike. 
 
 \paragraph{Identifying rewriting context:}
 Our next step is to identify the contextual examples where the novel feature belongs. One straightforward way to identify context is to leverage the trained segmentation model. For any new image, identify the same segment as the novel segment identified previously. Many contextual examples can be obtained in this way, by simply segmenting any non-novel bikes~(\ie, context samples) and picking the location of the part in the context samples corresponding to the novel part of the novel sample. As our goal is to transform non-novel bike designs into novel bike designs, we randomly select 5 of generated samples, hence likely common designs generated by the GAN, as the target context for rewriting. This allows the rewriting to make changes across different generated samples, which increases the probability of the novel features appearing in randomly generated samples. In the end, we apply rewriting using the novel feature and target context described above. Our overall approach is summarized in Fig.~\ref{fig:methodoverview}.

\section{EXPERIMENTAL SETTINGS}
In this section we describe the experimental approach taken to demonstrate the effectiveness of CreativeGAN for an example of novel bike design. Furthermore, we describe some of the implementation details and specifics of the methods applied in CreativeGAN.

\subsection{Bike Dataset}
\label{sec:data}
\begin{figure}[h]
\centering
\vskip -0.5in
\includegraphics[width=0.7\columnwidth]{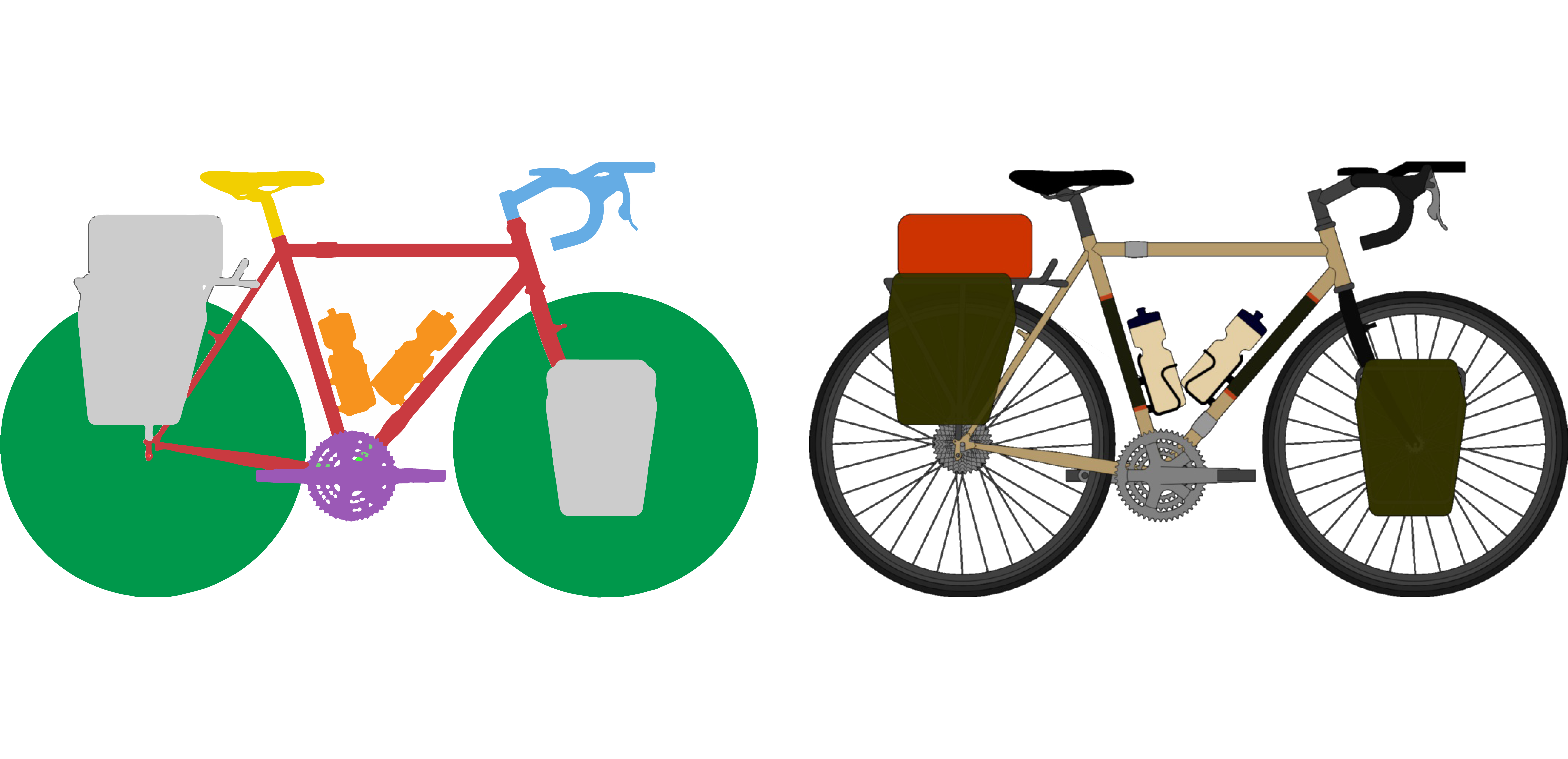}
\vskip -0.3in
\caption{Examples of the segmented bike images. Each color represent one part of the bike.}
\vskip -0.2in
\label{fig:segmentation_data}
\end{figure}

\vskip -0.1in
\begin{figure}[ht]
\centering
\vskip -0.2in
\includegraphics[width=\columnwidth]{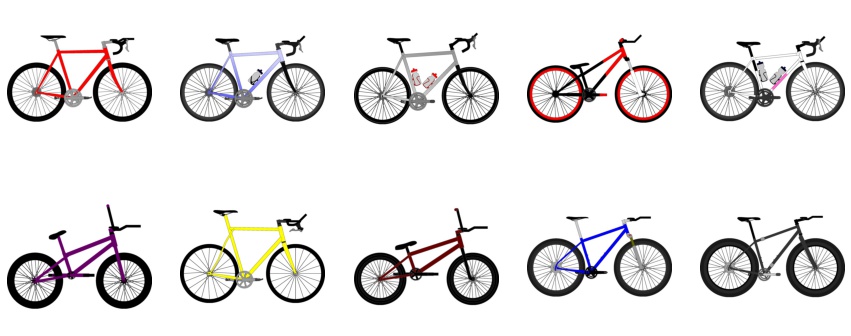}
\caption{Samples of the bike dataset.}
\label{fig:bike_dataset}
\vskip -0.2in
\end{figure}
In this paper we use a dataset of bikes, named BIKED, introduced by Regenwetter~\etal, which includes 4,775 bikes~\cite{bikedata} created by people on a bike design platform. The data are available in BikeCAD, CSV and image file formats. Fig.~\ref{fig:bike_dataset} shows a few sample images from the dataset. We use bike assembly and part images in our approach. The bike assembly images are used to train generative models, while we use the part images for training a segmentation model. To obtain images of each part of the bike separately, we edit the CAD files to include only one part of one bike at a time. Each bike is divided in seven main parts, which are frame, saddle, wheel, handle, bottle, rack, and crank. These bike part images are used to augment the data during training a GAN model to improve the quality of the generated samples. We also use these images to train a machine learning model to identify each part for a new bike by learning semantic segmentation information. A sample of the bike segmentation is shown in Fig.~\ref{fig:segmentation_data}. We then use segmentation information to train a deep convolutional neural network~(CNN)-based semantic segmentation model based on the U-Net architecture\cite{ronneberger2015unet}. After training our implementation of U-Net achieves an overall IoU score of 0.838, which indicates that the masks cover 83.8\% of the parts of the bike correctly. To avoid problems caused by the 16.2\% remaining error we dilate the masks for rewriting the GAN, which we describe in the following section.
\subsection{Implementation Details}
In our experiments, we use the novelty detection method to compute the anomaly scores of 1,000 generated samples and identify the top novel bikes. We use gray-scale images in the anomaly detection as we are more interested in the structure of the novel bikes than their color since we based our rewriting approach on bike parts. We select the 20 bikes with the highest novelty score~(Fig.~\ref{fig:noveltydetection_overall_both}) and determine the novel part of the bikes using the segmentation and novelty detection method described in the previous section. After this, we apply our rewriting method using the novel part of the bike as the input feature desired to be transferred to other samples. It is important to mention that to ensure complete coverage of the bike parts in our masks for rewriting, we dilate the masks using a dilation kernel size of 16px by 16px~(examples of these masks are illustrated in the leftmost images of Fig.~\ref{fig:rewriting_frame} and Fig.~\ref{fig:rewriting_handle}).
For the rewriting method, we found rewriting layer 6 to be the best qualitatively for rewriting the frame and layer 8 for rewriting the handle. In the optimization for rewriting, we use a learning rate of 0.05 optimize the model for 2000 steps and use a rank~\cite{bau2020rewriting} of 15 to edit the models.
\subsection{Evaluation Metrics}
To quantitatively measure if the editing applied by our method has guided the generator to synthesize novel designs, we employ the SPADE anomaly detection method to establish a metric to validate that the changes have resulted in novelty in the generated samples. To measure the novelty of a generator, we compute the average novelty score of 10,000 samples randomly generated by any edited generator. Beyond this we introduce a different metric to further validate our quantitative results. This time we use the images themselves rather than machine-learning based features. We measure novelty using the structural similarity index measure~(SSIM). To do this we measure the SSIM distance of every generated sample with all of the images in the dataset and use the the top 50 nearest neighbours as the measure for novelty. 
% To directly get novelty we turn the SSIM to what we call the SSIM distance, which measures a normalized distance based on the SSIM. SSIM normally ranges from -1 to 1. We normalize it by adding 1 to it and dividing by 2. Further we calculated the distance as the difference between this value and the maximum possible normalized similarity which is 1.0:
\begin{equation}
    SSIM~distance=\frac{1-SSIM}{2}
\end{equation}
In this metric the higher the SSIM distance the more novel a model is.
\section{RESULTS AND DISCUSSION}
We use the CreativeGAN method described above to guide the StyleGAN2 generator to generate novel designs. We showcase CreativeGAN through experiments on guiding novelty by editing the generator based on the novel frames and handles found in the top 20 most novel samples generated by the generator. Figure~\ref{fig:rewriting_frame} and~\ref{fig:rewriting_handle} demonstrate some of the experimental results showing how CreativeGAN changes the behavior of the original GAN.
\subsection{StyleGAN2 Results}
\begin{figure}[ht]
\centering
\vskip -0.2in
\includegraphics[width=\columnwidth]{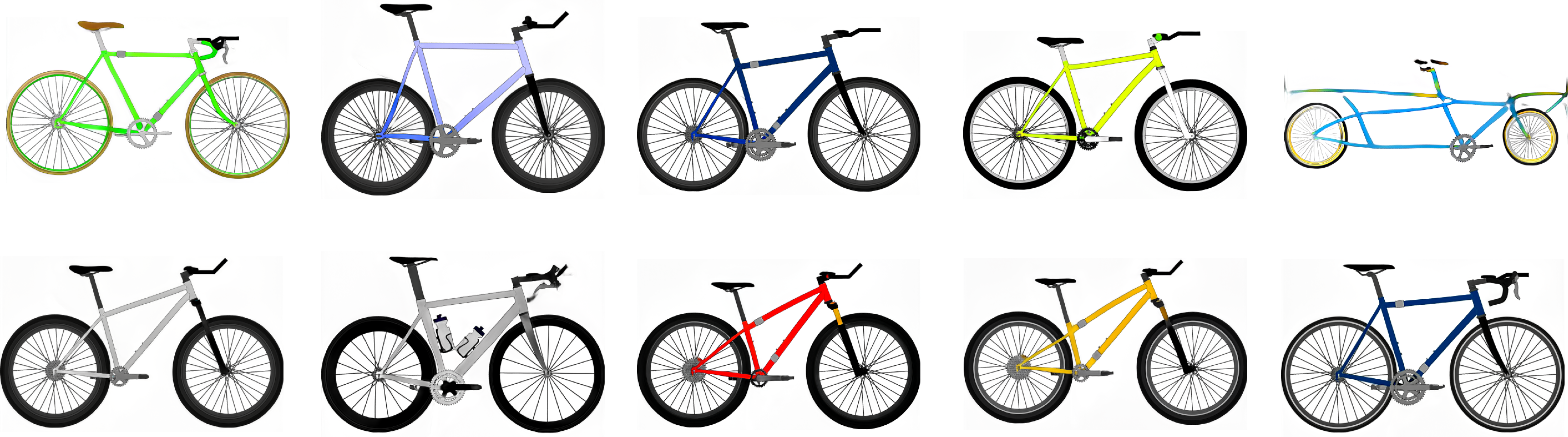}
\caption{Randomly generated samples of StyleGAN2 trained on bike images with our augmentation method.}
\label{fig:generated_samples}
\vskip -0.2in
\end{figure}

Figure~\ref{fig:generated_samples} demonstrates 8 randomly generated bikes by our model. Note that the GAN model can generate realistic-looking bikes, with consistency in color schemes within each model. It also does a good job at generating small parts, as seen by the presence of bottles and cranks. It also preserves symmetry in generated samples (the front and back wheels are of similar size). Overall, the bike synthesis results demonstrate that GAN models give good results in generating functional designs with multiple parts.
\begin{figure*}[ht!]
\centering
\includegraphics[width=1.9\columnwidth,height=2.4in]{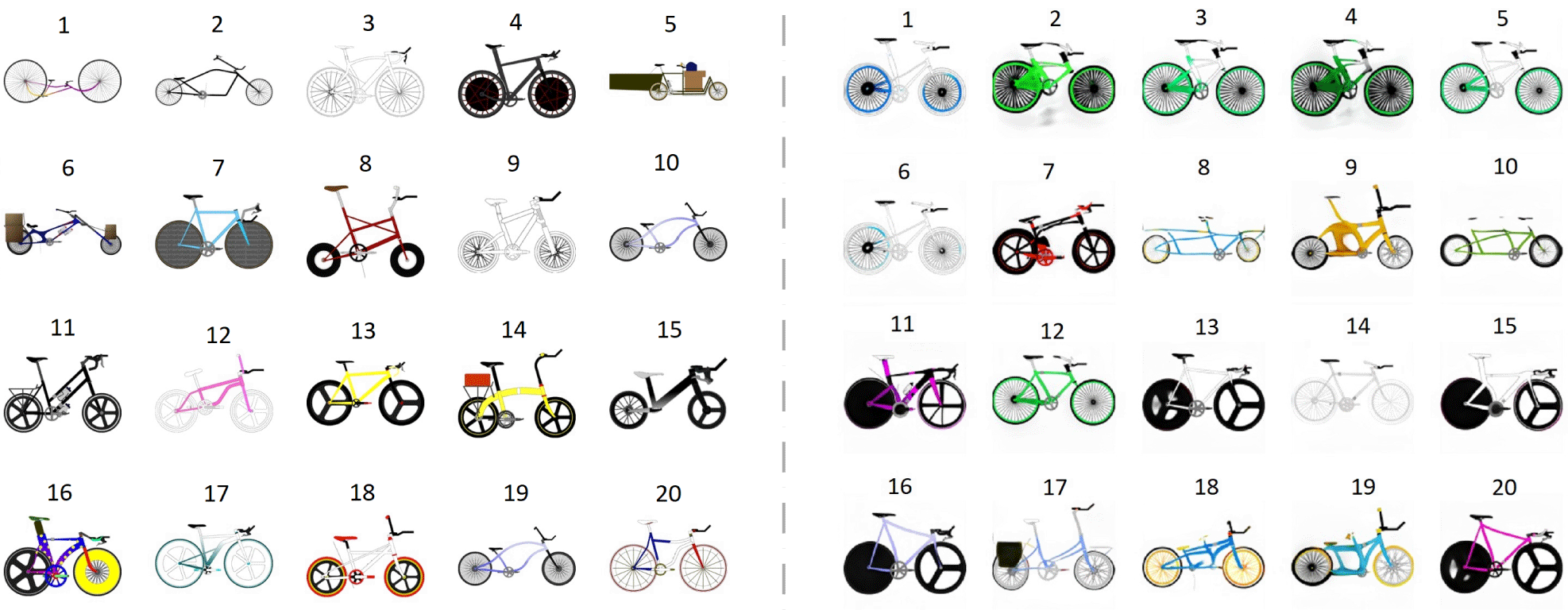}
\vskip -0.1in
\caption{Left: The top 20 most novel bikes in the BIKED dataset based on their novelty scores using our novelty detection approach. Right: The top 20 most novel bikes amongst 1,000 randomly generated samples based on their novelty scores using our novelty detection approach.}
\label{fig:noveltydetection_overall_both}
\end{figure*}
\subsection{Novelty Detection In Bikes}
To illustrate the efficacy of the novelty detection method employed in CreativeGAN we demonstrate the most novel designs from the training data as well as the designs synthesized by the GAN model. We first apply the novelty detection to the original bike dataset to identify unique designs in it. Fig.~\ref{fig:noveltydetection_overall_both}-Left depicts the top 20 most novel designs within the dataset ranked based on their novelty score. One can notice that the metric successfully identifies uncommon designs with unique frames, wheels or overall structure. Furthermore, we apply the same novelty scoring approach to 1,000 randomly generated designs by our GAN model and depict the top 20 most novel amongst them in Fig.~\ref{fig:noveltydetection_overall_both}-Right. It can be seen visually that the bikes ranked most novel in both often differ significantly in their design compared to the common samples seen in the dataset~(Fig.~\ref{fig:bike_dataset}), which contains 40\% road bikes.
\subsection{Introducing Novelty in Bikes}
\begin{figure*}[h!]
\centering
\includegraphics[width=1.7\columnwidth,height=3.0in]{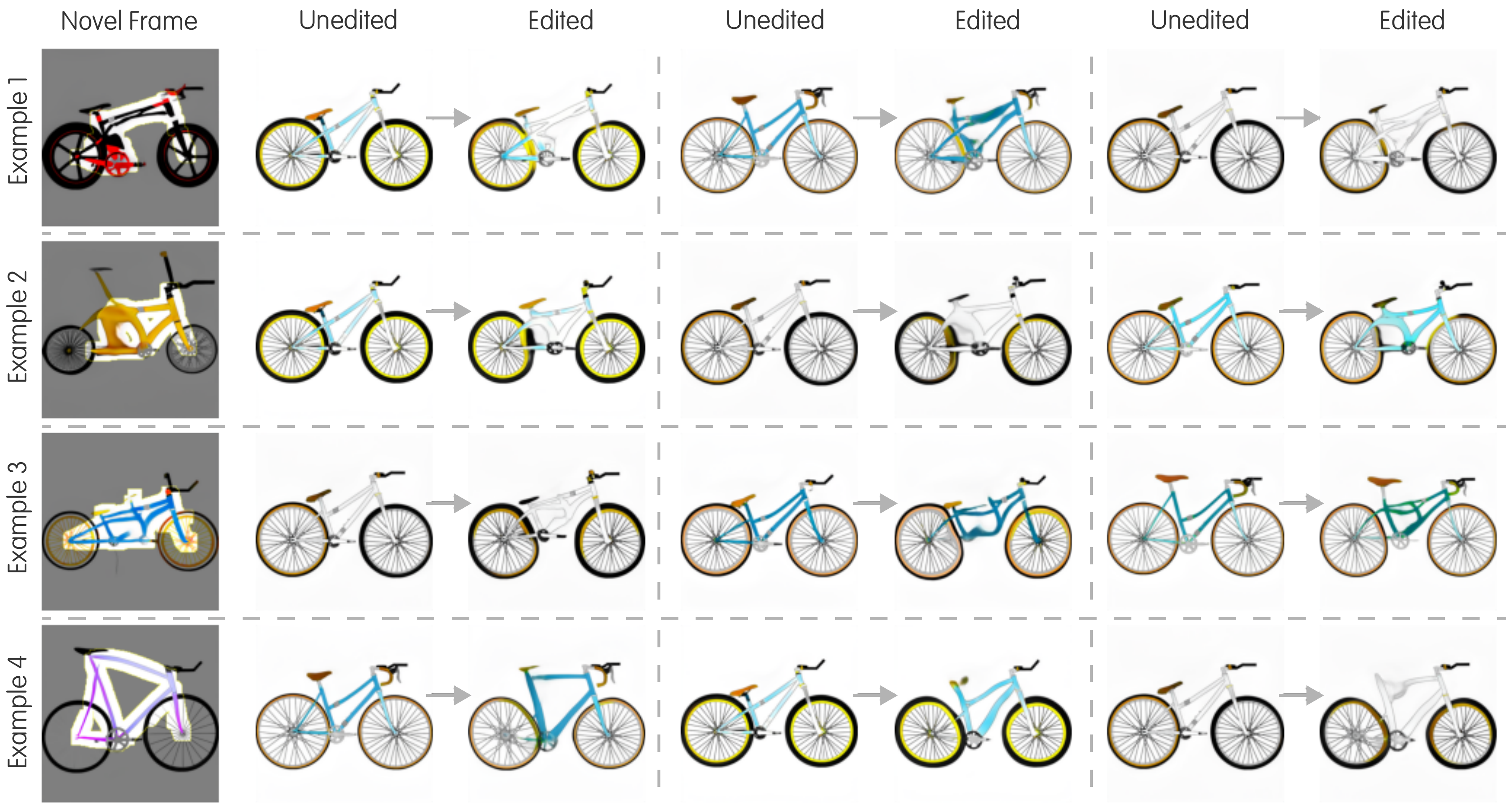}
\vskip 0.1in
\caption{Rewriting result on novel frame. The leftmost column shows the novel frames and the dilated segmentation mask used for rewriting. The right three columns each including pairs of images showing the transition of samples before and after CreativeGAN.}
\label{fig:rewriting_frame}
\end{figure*}
\begin{figure*}[h!]
\centering
\includegraphics[width=1.7\columnwidth,height=2.3in]{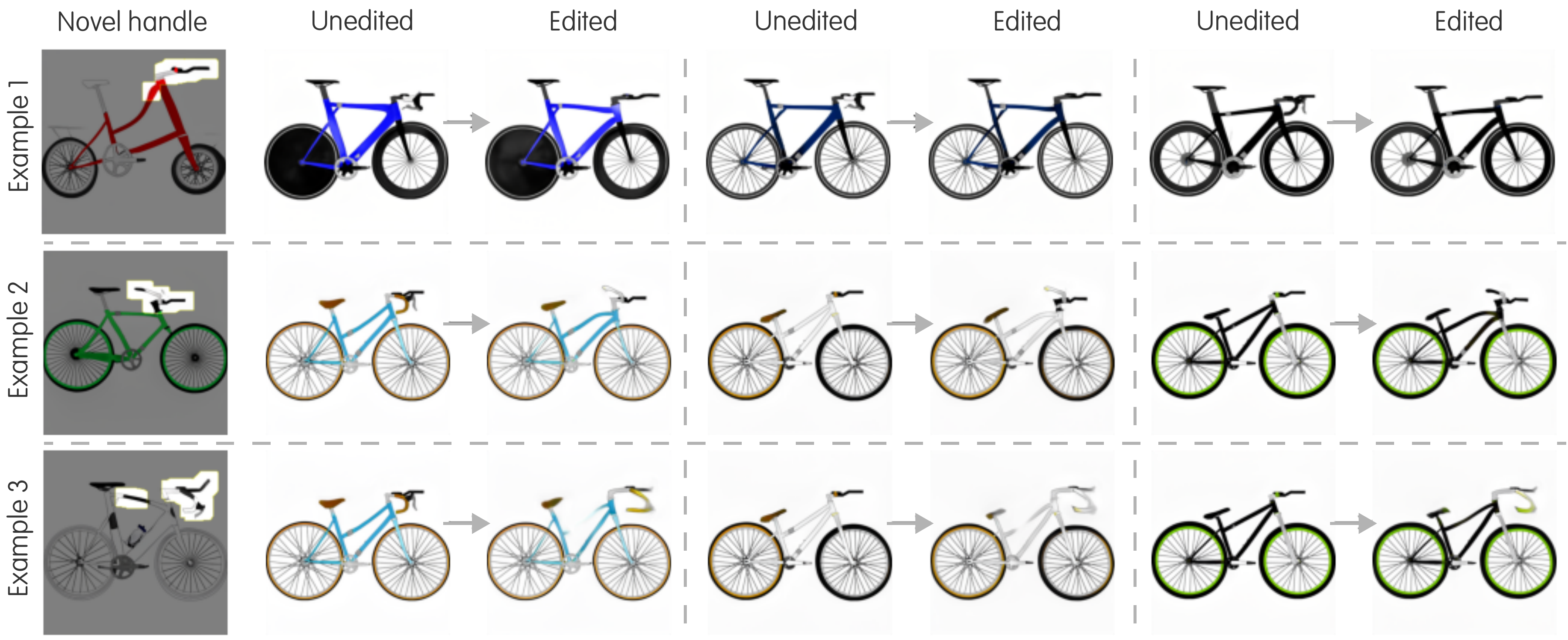}
\caption{Rewriting result on novel handles. The leftmost column shows the novel handle and the dilated segmentation mask used for rewriting. The right three columns each including pairs of images showing the transition of samples before and after CreativeGAN.}
\label{fig:rewriting_handle}
\end{figure*}
\begin{figure}[h!]
\centering
\includegraphics[width=\columnwidth]{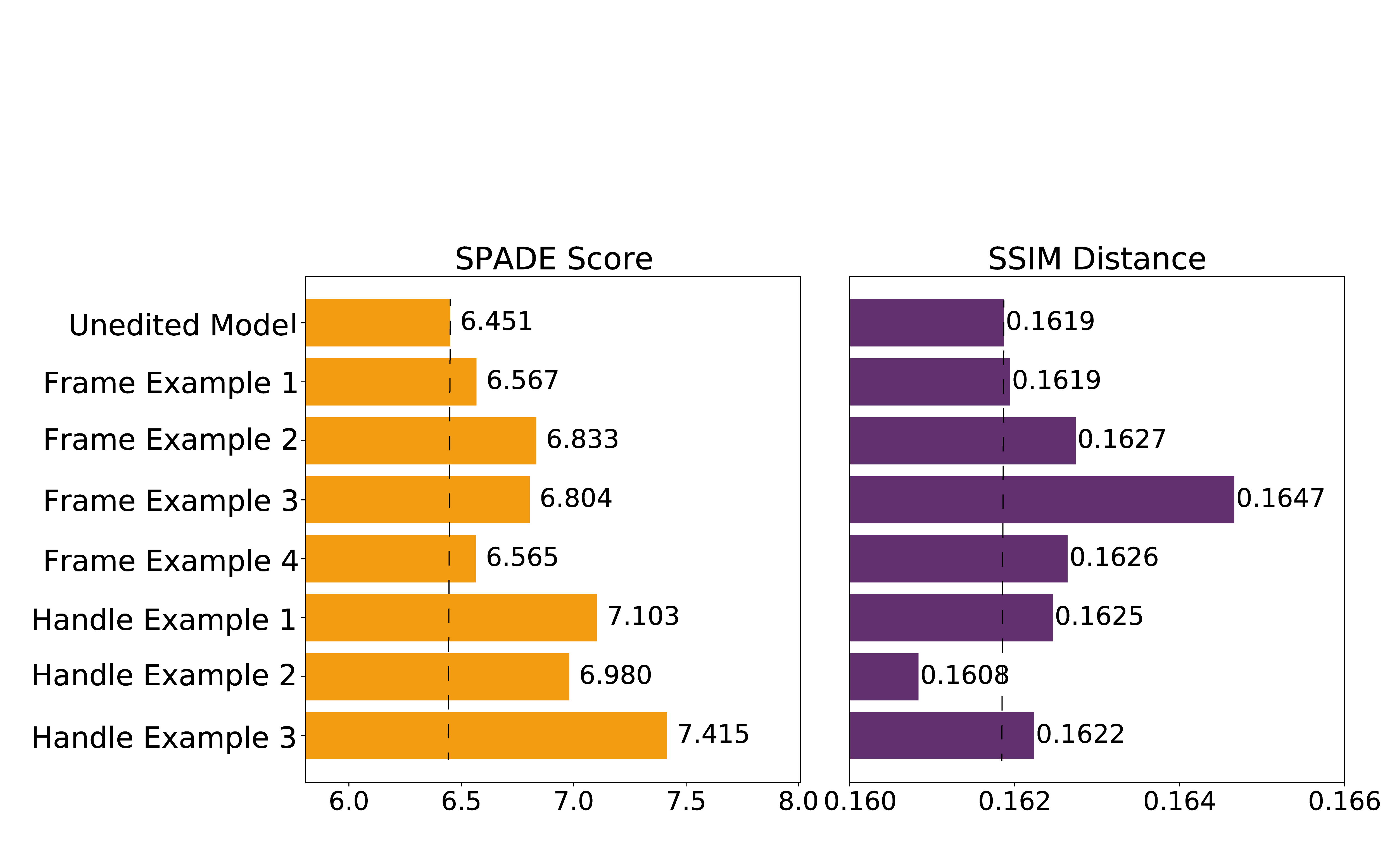}
\caption{Generator novelty scores in each experiment and the base-line unedited novelty of the StyleGAN2 generator.}
\vskip -0.2in
\label{fig:evaluatin_metric}
\end{figure}

In the experiments involving the novel bike frames~(Fig.~\ref{fig:rewriting_frame}) we observe that CreativeGAN is able to incorporate the structure of the novel features effectively in bikes. Of particular interest is CreativeGAN's ability to adapt unique frames from novel samples to common bikes with different saddles and handles. This is shown even if the saddle and handle have different locations, meaning the bikes have fundamentally different designs, CreativeGAN produces novel designs that do not change the fundamentals, but rather introduce the elements of novelty into those designs. This is significant from a design prospective as it shows that CreativeGAN does not simply replicate the novel design, but learns to generalize the novel elements. The results of introducing novel handles into the generator leads to realistic bike images as shown in Fig.~\ref{fig:rewriting_handle}. Introducing new handles seems to have worked as intended with little or no change to other parts of the bike design. This is expected, as bike handles have fewer connections to the rest of the bike design and do not effect other parts in any major way. The bikes produced by CreativeGAN in these examples do not exhibit distortions in other parts of the bike, indicating that the handles are disentangled from other parts of the bike in the generator. Further, We measure the average novelty score using the SPADE as well as the SSIM score for images generated by the GAN before and after applying CreativeGAN. The results of this are shown in Fig.~\ref{fig:evaluatin_metric}.
As evident across both metrics, CreativeGAN increases the novelty of samples notably, proving that CreativeGAN is able to modify GANs to generate novel designs. One exception here is in the SSIM distance of the handle example 2, which shows a decrease in the top 50 SSIM. As the SPADE novelty in this example is high in comparison to the unedited model, it is still possible that the overall novelty increased in this example as well, however, the lower SSIM distance does point to the possibility that the specific handle in example 2 may not be as novel as the SPADE method has detected it to be. Regardless, the visual inspection of novel designs and the agreement between SSIM and SPADE in six out of seven examples indicate that the SPADE novelty detection method is effective as an overall method of detecting novel samples. Furthermore, since the SPADE novelty metric is computed for 10,000 randomly generated samples compared to the 100 samples in SSIM~(given SSIM is computationally expensive), it demonstrate the overall novelty of the model better. The increase in this novelty metric demonstrates that CreativeGAN can enable a GAN to generate novel designs in a generalizable way, which effects the overall behaviour of the model.
% \begin{figure}[h!]
% \centering
% \includegraphics[width=0.8\columnwidth]{emb.png}
% \caption{Change in generated bike embeddings~(2 dimensional PCA) after applying CreativeGAN. This figure shows how CreativeGAN guides the generator towards more novel regions of the design space.}
% \vskip -0.2in
% \label{fig:bike_embedding}
% \end{figure}
% We further demonstrate the effect of CreativeGAN by computing the 2-dimensional embeddings of the generated samples and the dataset images using principal component analysis (PCA). We then show in Fig.~\ref{fig:bike_embedding} how CreativeGAN changes the distribution of data in this lower dimensional representation by plotting how some of the images evolve after applying CreativeGAN editing. We see that the samples move away from the densely populated common parts of the design space towards more sparse, hence novel regions of the design space. This further confirms CreativeGAN's effectiveness in promoting novelty.
\subsection{Limitations And Future Works}
CreativeGAN provides a new way to synthesize designs with unique components. However, we are still exploring the tip of the iceberg. It is important to discuss the limitations of CreativeGAN and the challenges that must be addressed for creative design generation. One issue that we observe is that when novel frames are drastically different, CreativeGAN fails to produce realistic designs. This issue is particularly notable in the bottom row (Example 4) of the results presented in Fig.~\ref{fig:rewriting_frame}, where a novel bike with a frame structure that is drastically disproportionate compared to the common bikes is identified for GAN editing, which leads to some generated samples having distorted frames or missing saddles. Another important observation is the fact that the wheels are often observed to be distorted slightly when CreativeGAN is trained on novel frames, the reason behind this is likely the fact that when generating designs the generator does not disentangle the frame from the wheels which has to lead to some distortion in the wheels when changes have been applied to the frames. This is a significant limitation of CreativeGAN and great effort must be made to ensure proper disentanglement between different parts of any design in the generator. In the future, we intend to develop approaches to improve the results of CreativeGAN and reduce the number of distortions and unwanted changes.

An important aspect of creativity is `usefulness' which we did not include in our modeling. Currently, CreativeGAN relies on the generator learning to produce useful designs based on the training dataset. However, there is no rule restricting the GAN to not produce useful designs.  In our model, we do not have a mechanism to promote the performance of bikes. In the future we intent to introduce mechanisms to promote generation of high performance bikes in the GAN. We also plan to introduce similar objectives to the constraint loss term in Eq.~\ref{eqn:2} to ensure editing is being done with performance and usefulness in mind, to improve creativity of components and entire design.

Finally, the novelty detection implemented here is not specialized for bikes and is a generalized method, which involves pre-trained models trained on ImageNET. We expect new insights if the novelty detection and localization were to be tailored specifically to any given application. In the future we intend to explore methods of novelty detection that are specifically tailored to CreativeGAN and GAN editing based on novelty and verify their efficacy using expert judgments. Furthermore, we intend to explore new design synthesis methods, which take into account multiple aspects of creativity such as novelty, usefulness, performance, cost, \etc{}.
\subsection{Broader Impacts of CreativeGAN In Design}
In this paper, we introduced a novel generalizable framework for promoting creativity in GANs. Automating creativity in design is a topic that is less explored in data-driven automated design synthesis. Due to a lack of common agreement on the definition of creativity, it is also less explored in the machine learning community working on generative models. Creativity in design is difficult to model and study and presents a limitation of real-world design applications of machine learning and data-driven design synthesis. Enabling machines to be creative has been a goal of artificial intelligence~(AI)-based automated design synthesis. At this point, most applications of AI and particularly machine learning in design focus primarily on mimicking existing designs or emulating designs that already exist in different ways. This although extremely valuable for design space exploration within the known realms of any field of design, provides no avenue for design space exploration beyond the existing known boundaries of the design space, and pioneering new designs that are truly creative and transformative. In this paper, we explorer an early proof of concept in automating creative design synthesis, showing that design space exploration in a truly creative and automated fashion may be possible using AI. Finally, it is important to note that the approaches introduced here are generalizable to many different fields of design, and can broadly be applied as a framework for promoting novelty in design synthesis using data-driven methods involving GANs. In this paper we specifically demonstrated our methods in a bike design example, however, the same approaches can be applied in other domains of design. Therefore, our contributions in this paper provide a stepping stone for other researchers to employ similar methodology to explore creativity in their own domains using GAN-based design synthesis.
\section{CONCLUSION}
In this paper, we introduce a novel approach, ``CreativeGAN'', for promoting creativity in GANs. We do this by combining principles of novelty detection using machine learning and modifying a pre-trained GAN, called `rewriting GANs'. By combining these two aspects we automate the process of guiding creativity in GANs without the need for human intervention, which combined with the automatic design synthesis of GANs automates the creativity process. We demonstrated that our method is capable of producing novel designs by detecting unique features within GAN-generated designs and applying these features in other bike designs to generate a large set of novel designs which incorporate the unique features discovered. In doing this the GAN generalizes these novelties into the design process and generates designs that are more novel. We verified the novelty of CreativeGAN generated designs both visually and through quantitative metrics and demonstrated that the approaches employed in this paper result in better novelty in generated samples. We discovered that CreativeGAN was able to adapt novel features in rare and novel bikes into more common bike designs without changing the overall structure of the common bikes, hence generating designs that integrate novelty in common designs to create previously undiscovered designs.
This paper provides a pathway for machine learning in design to think beyond interpolating existing designs to automate creative design synthesis and exploration. We show the potential for GAN-based approaches in going beyond design space exploration within the known boundaries of the design space and into pioneering transformative designs outside the known design space boundaries without human supervision.
\bibliographystyle{asmems4}

%%%%%%%%%%%%%%%%%%%%%%%%%%%%%%%%%%%%%%%%%%%%%%%%%%%%%%%%%%%%%%%%%%%%%%
\begin{acknowledgment}
The authors acknowledge the MIT SuperCloud and Lincoln Laboratory Supercomputing Center for providing HPC resources that have contributed to the research results reported within this paper.
\end{acknowledgment}

%%%%%%%%%%%%%%%%%%%%%%%%%%%%%%%%%%%%%%%%%%%%%%%%%%%%%%%%%%%%%%%%%%%%%%
% The bibliography is stored in an external database file
% in the BibTeX format (file_name.bib).  The bibliography is
% created by the following command and it will appear in this
% position in the document. You may, of course, create your
% own bibliography by using thebibliography environment as in
%
% \begin{thebibliography}{12}
% ...
% \bibitem{itemreference} D. E. Knudsen.
% {\em 1966 World Bnus Almanac.}
% {Permafrost Press, Novosibirsk.}
% ...
% \end{thebibliography}

% Here's where you specify the bibliography database file.
% The full file name of the bibliography database for this
% article is asme2e.bib. The name for your database is up
% to you.
\bibliography{asme2e}

\end{document}